\documentclass[runningheads]{llncs}

\usepackage[T1]{fontenc}
\usepackage{graphicx}
\usepackage{amsmath,amssymb}
\usepackage{subfig}
\usepackage{booktabs}
\usepackage{float}

\title{An Attention-Based Denoising Model for Diffusion Weighted Imaging}

\titlerunning{Attention-Based DWI Denoising}

\author{
Prithviraj Verma\inst{1} \and
Pawan Kumar\inst{1} \and
Chandan Deshani\inst{1} \and
Prasun Chandra Tripathi\inst{1,2}
}

\authorrunning{Verma et al.}

\institute{
Institute of Infrastructure Technology Research and Management (IITRAM), India \and
University of Sheffield, United Kingdom
}
\begin{document}

\maketitle

\begin{abstract}

Diffusion-weighted imaging (DWI) is used for whole-body cancer screening, but it typically requires a long acquisition time. When the scan time is reduced, the image quality often suffers, leading to increased noise in the scans. Magnitude reconstruction in DWI introduces signal-dependent Rician noise, which makes denoising more challenging for conventional convolution-based methods. To address this limitation, we propose a noise-aware attention-driven denoising framework that integrates hierarchical Swin Transformer window attention with transformer-based multi-dimensional gated refinement for DWI restoration. The model incorporates explicit noise-level conditioning and residual reconstruction to enable adaptive suppression of heteroscedastic noise across a wide range of corruption levels. Experimental evaluation on corrupted DWI scans demonstrates strong restoration performance. Our model achieves a mean PSNR of 33.69~dB and SSIM of 0.8539 across noise levels from 1\% to 15\%, while maintaining stable behavior under severe noise conditions. These results indicate that attention-guided contextual modeling combined with channel-adaptive refinement provides a robust and generalizable solution for DWI denoising.


\end{abstract}

\section{Introduction}

Diffusion-weighted Imaging (DWI) quantifies water molecule mobility for tissue microstructure assessment in stroke diagnosis, lesion detection, and white matter tractography. Echo-planar acquisition with high diffusion weighting produces intrinsically low signal-to-noise ratios, which degrades derived biomarkers such as apparent diffusion coefficient (ADC) and fractional anisotropy~\cite{magnotta2012multicenter}. In addition, magnitude reconstruction from complex receiver channels introduces signal-dependent Rician noise characterized by intensity bias in low-SNR regions and spatially heterogeneous variance, which differs fundamentally from additive Gaussian noise assumptions commonly used in natural images~\cite{wiest2008rician,coupe2008nonlocal}.

Rician statistics induce rectification bias where the noise floor elevates low-intensity signals, which leads to systematic distortion in quantitative diffusion measurements and potentially affecting diagnostic interpretation. The spatially varying nature of this corruption necessitates adaptive denoising strategies capable of preserving anatomical boundaries while accounting for local signal strength~\cite{coupe2008nonlocal}.

Classical denoising approaches including non-local means and block-matching filtering typically adapted for MRI rely on handcrafted priors and fixed similarity assumptions that limit their robustness under complex acquisition variability and increasing computational cost~\cite{wiest2008rician,coupe2008nonlocal}. Convolutional Neural Networks (CNNs) provide improved empirical performance through learned hierarchical feature representations~\cite{zhang2017dncnn,zhang2018ffdnet}, yet their locality-constrained receptive fields restrict effective contextual reasoning when homogeneous tissue regions are heavily corrupted. Encoder–decoder architectures such as U-Net remain widely adopted in medical imaging but still depend on progressive receptive-field expansion rather than explicit global dependency modeling~\cite{ronneberger2015unet}.

Recent transformer-based architectures~\cite{liang2021swinir,zamir2022restormer} address these limitations by enabling long-range spatial dependency modeling via self-attention. The Swin Transformer~\cite{liang2021swinir} introduces a hierarchical shifted-window attention mechanism that reduces computational complexity while still capturing global contextual relationships among image patches. Restormer architecture introduced in~\cite{zamir2022restormer} further extends restoration-specific transformer design through multi-dimensional transposed attention and gated depthwise convolution that enables adaptive channel–spatial feature refinement for image restoration tasks. More recent works have extended transformer-based denoising to medical imaging, demonstrating improved generalization and reconstruction fidelity in diffusion MRI and low-field MRI settings~\cite{sadikov2024generative,zhu2024imtmrd}.

Some works have also explored hybrid and noise-aware learning strategies for MRI denoising. In particular, transformer-CNN hybrid architectures~\cite{shi2025htcnet} with self-supervised pretraining have shown improved capability in modeling signal-dependent noise compared to conventional CNN-based approaches. Furthermore, self-supervised methods have been proposed to address the lack of clean ground-truth data in DWI.~\cite{pfaff2024dwi,wu2025difusion}.

Despite increasing interest in attention-driven medical image restoration, rigorous evaluation under realistic Rician noise conditions remains limited. Existing studies often rely on Gaussian noise assumptions or single-noise training regimes that do not reflect the heteroscedastic statistics of magnitude MRI~\cite{xie2020denoising,manjon2019mri} in clinical settings. Furthermore, while recent self-supervised and generative approaches show promising results, many lack explicit noise-level conditioning or do not fully exploit hierarchical attention mechanisms to capture local and global dependencies under Rician noise conditions~\cite{pfaff2024dwi,wu2025difusion}. Consequently, the robustness and generalization behavior of attention-based denoisers for DWI remain insufficiently characterized.

To address these challenges, this work proposes a noise-aware transformer-based denoising framework for DWI affected by signal-dependent Rician noise. The model integrates hierarchical self-attention with channel-adaptive gated refinement and explicit noise-level conditioning to enable adaptive suppression of heteroscedastic corruption across varying noise levels. This design supports structurally consistent restoration under both moderate and severe degradation conditions. The main contributions of this work are summarized as follows:

\begin{enumerate}

\item \textbf{Attention-based restoration for DWI:}  
A Swin Transformer–Restormer architecture combining hierarchical spatial self-attention with channel-adaptive gated refinement is developed for DWI denoising to enhance robustness and structural preservation relative to convolutional baselines.

\item \textbf{Noise-aware denoising:}  
A noise-conditioned residual learning paradigm is employed to train the proposed model across multiple Rician noise levels (1\%--15\%) to allow adaptive denoising behavior and systematic evaluation of cross-regime generalization.

\item \textbf{Controlled comparison under Rician noise:}  
A rigorously controlled experimental framework is introduced to isolate architectural effects in DWI denoising under signal-dependent Rician corruption to enable fair comparison between convolution-only and attention-based restoration strategies.

\end{enumerate}

\section{Related Work}

\subsection{CNN-Based Image Denoising}

Convolutional neural networks (CNNs) have demonstrated strong effectiveness in image denoising through hierarchical feature learning and residual reconstruction~\cite{tripathi2020cnn}. As summarized in Table~\ref{tab:related_comparison}, DnCNN~\cite{zhang2017dncnn} showed that residual learning enables suppression of additive Gaussian noise across multiple corruption levels, while FFDNet~\cite{zhang2018ffdnet} introduced explicit noise-level conditioning to improve flexibility and generalization. However, these approaches assume signal-independent Gaussian noise and therefore do not directly address the signal-dependent corruption present in magnitude MRI~\cite{mohan2014survey}. Encoder--decoder architectures such as U-Net provide multi-scale contextual aggregation through skip connections~\cite{ronneberger2015unet,tripathi2020cnn}, yet their convolution-only receptive fields remain inherently local, limiting effective global structural reasoning in severely degraded DWI.

\subsection{MRI Denoising and Rician Noise}

Magnitude MRI noise follows Rician or non-central Chi statistics to produce intensity bias in low-signal regions and spatially heterogeneous variance~\cite{wiest2008rician,coupe2008nonlocal}. Classical denoising methods such as non-local means improve structural preservation but rely on handcrafted similarity priors and incur high computational cost~\cite{coupe2008nonlocal}. There are trade-offs between statistical fidelity, edge preservation and computational efficiency across spatial, transform and learning-based MRI denoising paradigms~\cite{mohan2014survey}. Data-driven training strategies such as Noise2Noise~\cite{lehtinen2018noise2noise} enable learning without clean targets, yet underlying convolutional architectures typically lack explicit modeling of heterogeneous noise severity in DWI.

\subsection{Transformer-Based Image Restoration}

Self-attention enables direct long-range spatial dependency modeling. In~\cite{liang2021swinir}, SwinIR introduced hierarchical shifted-window attention for efficient global contextual reasoning. On the other hand, Restormer architecture developed in ~\cite{zamir2022restormer} presents restoration-specific transformer design by exploiting multi-dimensional transposed attention and gated depthwise refinement. Despite promising results in natural images, applications to DWI under realistic Rician noise remain limited, with many studies relying on Gaussian assumptions or single-noise training regimes~\cite{xie2020denoising,manjon2019mri}.

To address these gaps, we propose a noise-aware Swin--Restormer denoising framework adapted for DWI. The proposed model integrates hierarchical contextual attention, channel-adaptive refinement, and explicit noise-level conditioning within a residual reconstruction framework.

\begin{table}[!t]
\centering
\caption{Representative denoising methods and characteristics.}
\label{tab:related_comparison}

\begin{tabular}{l c c c}
\hline
Method & Type & Noise Model & Noise-Aware \\
\hline
DnCNN~\cite{zhang2017dncnn} & CNN & Gaussian & No \\
FFDNet~\cite{zhang2018ffdnet} & CNN & Gaussian & Yes \\
U-Net~\cite{ronneberger2015unet} & CNN & General & No \\
NLM-Rician~\cite{coupe2008nonlocal} & Classical & Rician & No \\
SwinIR~\cite{liang2021swinir} & Transformer & Gaussian & No \\
Restormer~\cite{zamir2022restormer} & Transformer & Gaussian & No \\
\hline
\end{tabular}

\end{table}

\section{Methodology}
\label{sec:method}

\subsection{The Proposed Swin--Restormer Architecture}

\begin{figure*}
\centering
\includegraphics[width=\textwidth]{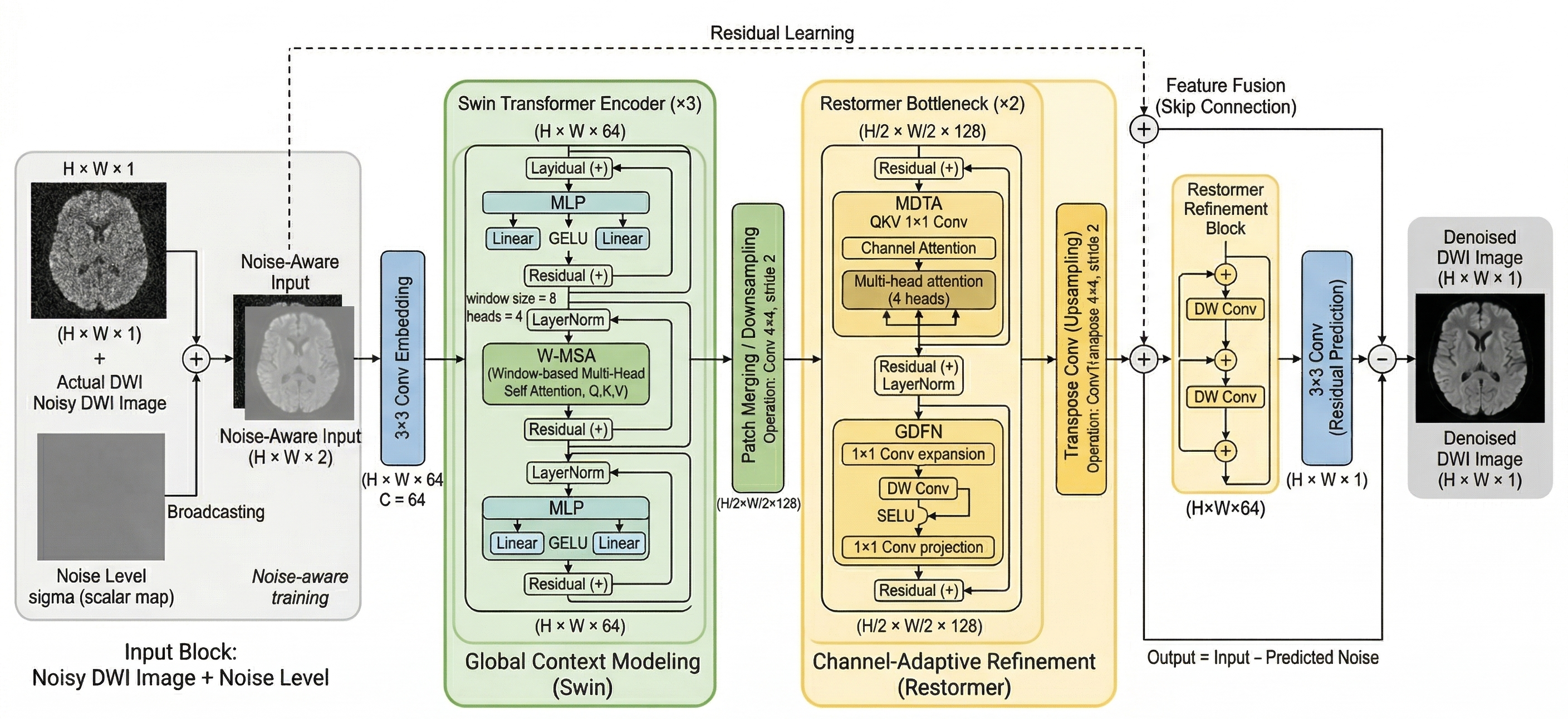}
\caption{The proposed Swin--Restormer architecture showing encoder, bottleneck and decoder stages with explicit noise-level conditioning. Here, $H$ and $W$ denote the spatial height and width of the input and output images. MDTA represents  multi-dimensional transposed attention and GDFN is Gated depthwise convolutional feed-forward networks.}
\label{fig:architecture}
\end{figure*}

Figure~\ref{fig:architecture} illustrates the proposed denoising network which integrates hierarchical window-based self-attention with multi-dimensional channel-adaptive refinement across encoder, bottleneck, and decoder stages. The design combines contextual modeling capabilities of the Swin Transformer~\cite{liang2021swinir} with restoration-oriented attention mechanisms to enable robust suppression of signal-dependent Rician noise in DWI scans. This design enables complementary modeling of long-range spatial dependencies and channel-wise feature refinement, which is particularly beneficial for suppressing signal-dependent noise. Below, we describe building blocks of our model.

\noindent\textbf{Embedding:}  
An initial $3 \times 3$ convolution layer projects the concatenation of the noisy diffusion-weighted image and spatially broadcast noise-level map into a 64-dimensional feature space for early fusion of corruption magnitude information.

\noindent\textbf{Window Attention Encoding:}  
Three Swin Transformer blocks perform shifted-window multi-head self-attention over non-overlapping $8 \times 8$ regions. Alternating shifted partitioning enables cross-window communication, to set effective global receptive fields while reducing computational complexity from $\mathcal{O}((HW)^2)$ to $\mathcal{O}(HW \cdot M^2)$~\cite{liang2021swinir}.

\noindent\textbf{Bottleneck Processing:}  
Following spatial downsampling, two Restormer blocks apply multi-dimensional transposed attention (MDTA) to model spatial and channel relationships for enabling adaptive feature selection for restoration~\cite{zamir2022restormer}. Gated depthwise convolutional feed-forward networks (GDFN) further enhance nonlinear transformation efficiency.

\noindent\textbf{Residual Reconstruction:}  
Transposed convolution restores spatial resolution, followed by additional Restormer refinement and a final $1 \times 1$ convolution predicting the noise residual $\hat{N}$. The denoised image $\hat{X}$ is reconstructed via residual learning as

\begin{equation}
\hat{X} = Y - \hat{N},
\end{equation}
\noindent
Here, $Y$ is the noisy input, $\hat{N}$ the predicted noise, $\hat{X}$ the denoised output and $X$ the clean reference image. This strategy is widely adopted in modern denoising networks for improved convergence and detail preservation~\cite{zhang2017dncnn}. The height and width of images $\hat{X}$ and $Y$ are $H$ and $W$, respectively.

\subsection{Swin and Restormer Mechanisms}

Given feature map $\mathbf{F} \in \mathbb{R}^{H \times W \times C}$, Swin Transformer partitions features into $M \times M$ windows and computes self-attention using the following equation:

\begin{equation}
\mathrm{Attention}(\mathbf{Q}, \mathbf{K}, \mathbf{V})
=
\mathrm{Softmax}\!\left(\frac{\mathbf{Q}\mathbf{K}^{T}}{\sqrt{d_k}} + \mathbf{B}\right)\mathbf{V},
\end{equation}
\noindent
where $\mathbf{Q},\mathbf{K},\mathbf{V}$ are learned projections, $d_k$ is the key dimension and $\mathbf{B}$ denotes relative positional bias~\cite{liang2021swinir}. Shifted window partitioning enables progressive global information exchange across layers. Restormer blocks employ multi-dimensional transposed attention operating across channel representations allowing adaptive channel reweighting based on restoration relevance during denoising~\cite{zamir2022restormer}. The attention-enhanced feature map $\mathbf{F}_{\mathrm{att}}$ is obtained via channel-wise self-attention using the following equation:

\begin{equation}
\mathbf{F}_{\mathrm{att}} = \sigma\!\left(\mathbf{Q}_c \mathbf{K}_c^{T}\right)\mathbf{V}_c,
\end{equation}
\noindent
The query, key, and value projections are represented by $\mathbf{Q}_c$, $\mathbf{K}_c$, and $\mathbf{V}_c$, respectively. The attention map is computed by taking the matrix multiplication of $\mathbf{Q}_c$ and the transpose of $\mathbf{K}_c$, followed by softmax function $\sigma(\cdot)$ to capture inter-channel dependencies. This attention map is then applied to $\mathbf{V}_c$ to selectively emphasize informative features while suppressing less relevant ones, enabling more effective feature refinement. The gated depthwise convolutional feed-forward module produces $\mathbf{F}_{\mathrm{out}}$ is expressed as

\begin{equation}
\mathbf{F}_{\mathrm{out}} =
\phi\!\left(W_2 \cdot \mathrm{DWConv}(W_1 \mathbf{F})\right)
\odot
\left(W_2 \cdot \mathrm{DWConv}(W_1 \mathbf{F})\right),
\end{equation}
\noindent
where $\phi$ denotes GELU activation, $\odot$ element-wise multiplication, and DWConv is depthwise convolution.

\subsection{Noise-Aware Conditioning}

The Rician noise standard deviation $\sigma$ is encoded as a scalar and spatially broadcast to form a single-channel noise map concatenated with the noisy input prior to feature embedding. During training, $\sigma$ is uniformly sampled from eight predefined corruption levels (1\%--15\%), encouraging the network to learn noise-adaptive representations rather than single-regime denoising. Explicit noise conditioning has been shown to improve robustness and generalization in CNN-based denoisers~\cite{zhang2018ffdnet}.

\begin{figure*}[!t]
\centering
\includegraphics[width=\textwidth]{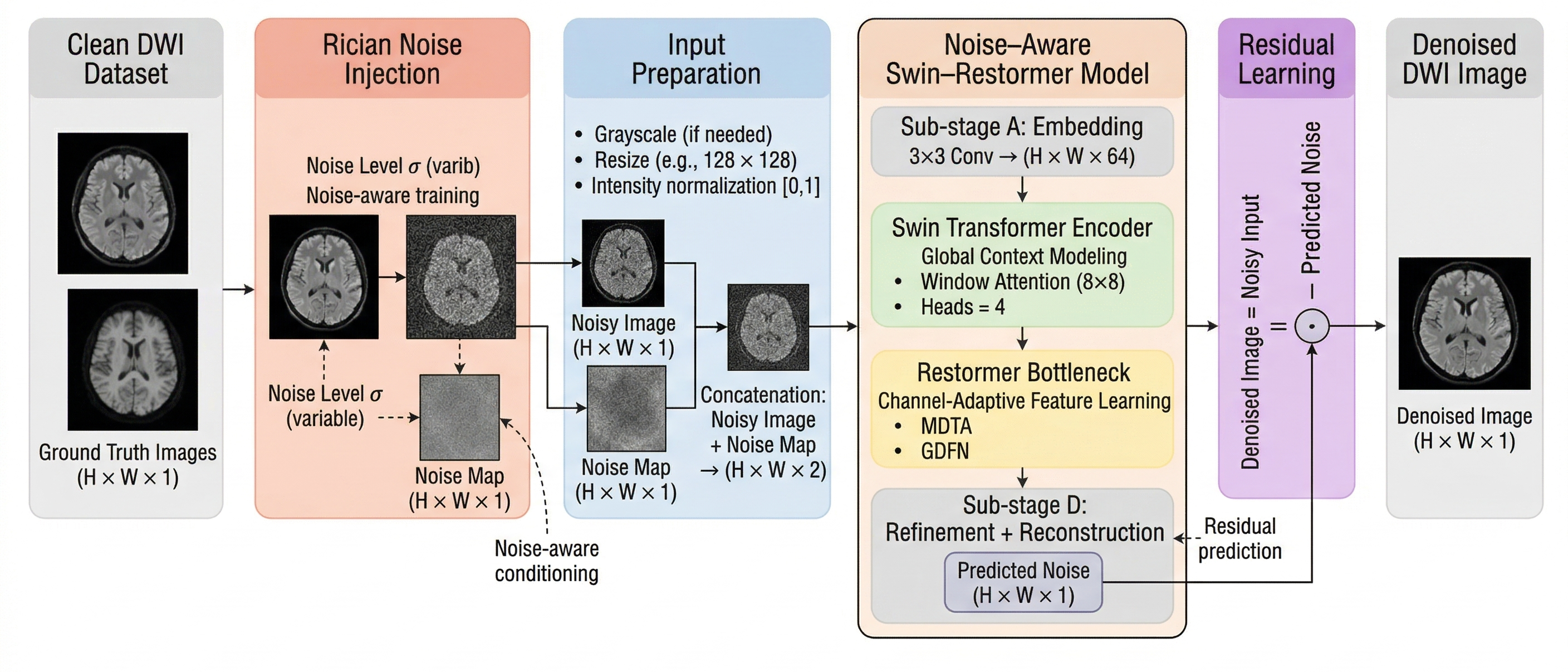}
\caption{Complete denoising workflow from preprocessing through residual reconstruction.}
\label{fig:workflow}
\end{figure*}
\subsection{Workflow and Training Objective}

Figure~\ref{fig:workflow} presents the full denoising pipeline:  
(a) DWI normalization to $[0,1]$;  
(b) synthetic Rician noise corruption;  
(c) noise-level encoding and concatenation;  
(d) transformer-based encoding, bottleneck refinement and decoding;  
(e) residual subtraction producing the final restored image. We minimize a hybrid Charbonnier–SSIM loss function $\mathcal{L}$ during training, which is expressed as follows:

\begin{equation}
\mathcal{L}
=
\sqrt{\|\hat{X} - X\|^2 + 10^{-6}}
+
\alpha \bigl(1 - \mathrm{SSIM}(\hat{X}, X)\bigr),
\end{equation}
\noindent
where $\alpha$ controls the relative contribution of structural similarity and is set to $\alpha = 0.3$ in all experiments in this work. The loss function balances pixel-level fidelity with structural similarity, a strategy commonly used in medical image restoration tasks to preserve perceptual and anatomical consistency.

\section{Experimental Results and Analysis}
\label{sec:setup}

\subsection{Dataset}

We utilized diffusion-weighted imaging data from the Traveling Human Phantom (THP) study which is a multi-site reliability dataset containing repeated acquisitions of five healthy subjects scanned across eight imaging centers~\cite{magnotta2012multicenter}. The dataset includes diffusion-weighted, T1-weighted MP-RAGE and T2-weighted SPACE sequences, enabling analysis of intra-subject and inter-site variability in diffusion-derived metrics. For this work, the processed 2D DWI slices yield 25,200 samples. All images were resized to $128 \times 128$ and normalized to $[0,1]$. Random $64 \times 64$ crops were extracted during training to increase diversity and mitigate anatomical overfitting. Data was partitioned into 70\% training, 10\% validation and 20\% testing splits.

\subsection{Training Configuration}

We used the Adam optimizer with $\beta_1=0.9$ and $\beta_2=0.999$.  
The initial learning rate was fixed to $5 \times 10^{-4}$ with cosine annealing scheduling. We trained the model for 30 epochs using batch size 4. We used $64 \times 64$ patches for training. In our model, we employed 64 feature channels, four attention heads and window size 8. Noise-aware training samples were corrupted uniformly from 1\%–15\%. The performance was measured using PSNR and SSIM across all noise levels. All models were implemented using the PyTorch deep learning framework and trained on a workstation equipped with an NVIDIA RTX 5050 GPU, an Intel Core i7 240H processor (2.50 GHz) and 24 GB of system memory. Python version $3.11$ was used for the experimentation.

\subsection{Noise Modeling}

Synthetic Rician corruption follows MRI magnitude reconstruction physics as

\begin{equation}
Y = \sqrt{(X + n_1)^2 + n_2^2},
\end{equation}
\noindent
where $X$ denotes the underlying clean (noise-free) image, $Y$ represents the corresponding observed noisy magnitude image, and $n_1,n_2 \sim \mathcal{N}(0,\sigma^2)$ are independent zero-mean Gaussian noise components~\cite{wiest2008rician}. The resulting magnitude signal exhibits signal-dependent bias and heteroscedastic variance that are characteristic of Rician noise.

Noise severity is defined by $\sigma$ as a percentage of normalized intensity, with eight levels from 1\% to 15\%. Low-SNR regions approximate Rayleigh statistics while high-SNR regions approach Gaussian behavior, enabling systematic evaluation of denoising robustness under realistic DWI noise conditions~\cite{coupe2008nonlocal}.

\subsection{Experimental Results}

\begin{figure*}
\centering

\subfloat[Clean]{\includegraphics[width=0.3\textwidth]{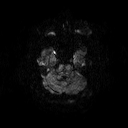}}
\hspace{2mm}
\subfloat[Noisy]{\includegraphics[width=0.3\textwidth]{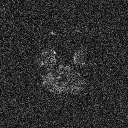}}
\hspace{2mm}
\subfloat[Denoised]{\includegraphics[width=0.3\textwidth]{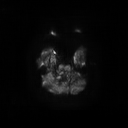}}

\vspace{2mm}

\subfloat[Clean]{\includegraphics[width=0.3\textwidth]{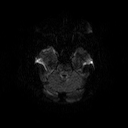}}
\hspace{2mm}
\subfloat[Noisy]{\includegraphics[width=0.3\textwidth]{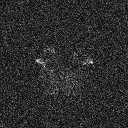}}
\hspace{2mm}
\subfloat[Denoised]{\includegraphics[width=0.3\textwidth]{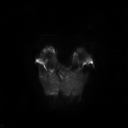}}

\vspace{2mm}

\subfloat[Clean]{\includegraphics[width=0.3\textwidth]{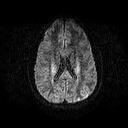}}
\hspace{2mm}
\subfloat[Noisy]{\includegraphics[width=0.3\textwidth]{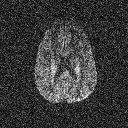}}
\hspace{2mm}
\subfloat[Denoised]{\includegraphics[width=0.3\textwidth]{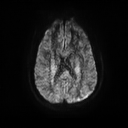}}

\caption{Qualitative DWI denoising results on three scans at 13\% Rician noise level.}
\label{fig:qualitative_results}

\end{figure*}

\noindent
Figure.~\ref{fig:qualitative_results} illustrates qualitative denoising results under 13\% Rician noise. The noisy images suffer from significant degradation and structural distortion, while the restored outputs exhibit effective noise suppression and improved preservation of fine anatomical details. Notably, the proposed method maintains edge sharpness and structural consistency across samples, indicating its robustness under high corruption levels. 

The quantitative denoising results on the test set for noise levels in the range [1, 15] are presented in Table~\ref{tab:noise_performance}. As expected, the performance of the models decreases with increasing noise levels. However, the proposed model continues to perform well even at higher noise levels (i.e., $13\%$ and $15\%$), which demonstres its strong generalization capability across varying noise conditions. We further compare our model with existing methods in Table~\ref{tab:literature_comparison}. The comparison is based on the mean performance across different noise levels. As shown in the table, the proposed model consistently performs better than the existing methods, which indicates its effectiveness and reliability for DWI denoising.

\begin{table}[!t]
\centering
\caption{The performance of the proposed model across different Rician noise levels.}
\label{tab:noise_performance}

\begin{tabular}{c c c}
\hline
Noise Level (\%) & PSNR (dB) & SSIM \\
\hline
1  & 43.82 & 0.9740 \\
3  & 37.12 & 0.9030 \\
5  & 34.43 & 0.8610 \\
7  & 32.76 & 0.8316 \\
9  & 31.58 & 0.8081 \\
11 & 30.67 & 0.7873 \\
13 & 29.91 & 0.7666 \\
15 & 29.24 & 0.7391 \\
\hline
\end{tabular}

\end{table}

\begin{table}
\centering
\caption{Quantitative comparison of the proposed model with existing methods.}
\label{tab:literature_comparison}

\begin{tabular}{l l l l c c}

\hline
Author & Method & Noise Type & Dataset & PSNR & SSIM \\
\hline
Wiest-Daesslé et al.~\cite{wiest2008rician} & NLM-Rician & Rician & DW-MRI & 27.6 & 0.81 \\
Coupé et al.~\cite{coupe2008nonlocal} & Optimized NLM & Rician & 3D Brain MRI & 28.4 & 0.83 \\
Manjón et al.~\cite{manjon2019mri} & NLPCA & Rician & Brainweb MRI & 31.2 & 0.88 \\
Xie et al.~\cite{xie2020denoising} & U-Net & Gaussian/Rician & ASL MRI & 26.3 & 0.79 \\
Xie et al.~\cite{xie2020denoising} & DilatedNet & Gaussian/Rician & ASL MRI & 27.8 & 0.82 \\
Xie et al.~\cite{xie2020denoising} & DWAN & Gaussian/Rician & ASL MRI & 29.4 & 0.85 \\
Zhang et al.~\cite{zhang2017dncnn} & DnCNN & Gaussian & Brain MRI & 28.5 & 0.82 \\
Zhang et al.~\cite{zhang2018ffdnet} & FFDNet & Gaussian & Brain MRI & 29.2 & 0.84 \\
\hline
UNet Baseline~\cite{ronneberger2015unet} & UNet & Rician & THP DWI & 24.27 & 0.5518 \\
\textbf{Proposed Model} & \textbf{Swin--Restormer} & \textbf{Rician} & \textbf{THP DWI} & \textbf{33.69} & \textbf{0.8539} \\
\hline

\end{tabular}

\end{table}

\subsection{Ablation Study}
\label{subsec:ablation}

To better understand the contribution of each architectural component, we conducted an ablation study evaluating different variants of the proposed denoising framework on the THP dataset. Specifically, we analyzed the impact of hierarchical Swin Transformer attention and Restormer-based gated refinement modules both independently and in combination.

Four model configurations were evaluated:

\begin{enumerate}

\item \textbf{UNet Baseline:} A convolutional encoder–decoder architecture trained under identical Rician noise conditions.

\item \textbf{Swin Transformer Only:} A denoising model using hierarchical Swin attention blocks without Restormer refinement.

\item \textbf{Restormer Only:} A transformer-based restoration model utilizing multi-dimensional transposed attention and gated depthwise convolution without Swin window attention.

\item \textbf{Proposed Swin--Restormer Model:} The complete architecture combining hierarchical Swin attention, Restormer refinement modules and explicit noise-aware conditioning.

\end{enumerate}

\begin{table}[t]
\centering
\caption{Ablation study evaluating the contribution of architectural components in the proposed denoising framework.}
\label{tab:ablation}

\begin{tabular}{lcc}

\hline
Model Variant & PSNR (dB) & SSIM \\
\hline

UNet Baseline & 26.27 & 0.6518 \\

Swin Transformer Only & 31.61 & 0.8489 \\

Restormer Only & 31.83 & 0.8347 \\

\textbf{Proposed Swin--Restormer} & \textbf{33.69} & \textbf{0.8539} \\

\hline

\end{tabular}

\end{table}
All model variants were trained from scratch on the THP dataset using identical training configurations, noise models and optimization settings to ensure a fair comparison. Table~\ref{tab:ablation} reports the results of ablation study. These results demonstrate the complementary strengths of the two attention mechanisms. The Swin-only configuration significantly improves performance over the convolutional UNet baseline by introducing hierarchical window-based attention capable of modeling long-range spatial dependencies. The Restormer-only architecture also provides strong restoration performance through channel-adaptive multi-dimensional attention and gated depthwise convolution.

The proposed Swin--Restormer architecture achieves the highest reconstruction quality, reaching a PSNR of 33.69~dB while maintaining strong structural similarity. These results indicate that hierarchical spatial attention and channel-adaptive refinement provide complementary benefits when combined this enables more effective suppression of signal-dependent Rician noise while preserving anatomical structures in diffusion-weighted imaging.

\section{Conclusion}

This work proposes a noise-aware Swin--Restormer framework for DWI denoising under signal-dependent Rician noise. The proposed architecture integrates long-range spatial attention with adaptive channel-wise refinement to achieve stable structural preservation across diverse noise conditions. The proposed model achieves robust restoration performance across a wide range of corruption levels, achieving a mean PSNR of 33.69~dB and SSIM of 0.8539 on THP DWI dataset while maintaining stable behavior under severe noise conditions. These results confirms the importance of attention-driven contextual modeling and heteroscedastic noise awareness for medical image restoration.

\section*{Acknowledgement}

The authors would like to acknowledge the support received under the project ANRF/ECRG/2024/005934/ENS. The research work presented in this paper was carried out with the support of this grant. The authors sincerely thank the funding agency for enabling this research.

\bibliographystyle{splncs04}
\bibliography{refs}

\end{document}